\documentclass[10pt,twocolumn,letterpaper]{article}

\usepackage{cvpr}              
\definecolor{cvprblue}{rgb}{0.21,0.49,0.74}
\usepackage[pagebackref,breaklinks,colorlinks,allcolors=cvprblue]{hyperref}

\def\paperID{*****} 
\def\confName{CVPR}
\def\confYear{2026}

\title{Fair Lung Disease Diagnosis from Chest CT via Gender-Adversarial Attention Multiple Instance Learning}

\author{
Aditya Parikh \quad Aasa Feragen \\
Technical University of Denmark (DTU)\\
Kongens Lyngby, Denmark\\
{\tt\small adipa@dtu.dk}
}

\begin{document}
\maketitle
\begin{abstract}
We present a fairness-aware framework for multi-class lung disease
diagnosis from chest CT volumes, developed for the Fair Disease Diagnosis
Challenge at the PHAROS-AIF-MIH Workshop (CVPR 2026).
The challenge requires classifying CT scans into four categories -- Healthy,
COVID-19, Adenocarcinoma, and Squamous Cell Carcinoma -- with performance
measured as the average of per-gender macro F1 scores, explicitly
penalizing gender-inequitable predictions.
Our approach addresses two core difficulties: the sparse pathological signal
across hundreds of slices, and a severe demographic imbalance compounded
across disease class and gender.
We propose an attention-based Multiple Instance Learning (MIL) model on a
ConvNeXt backbone that learns to identify diagnostically relevant slices
without slice-level supervision, augmented with a Gradient Reversal Layer
(GRL) that adversarially suppresses gender-predictive structure in the
learned scan representation.
Training incorporates focal loss with label smoothing, stratified
cross-validation over joint (class, gender) strata, and targeted
oversampling of the most underrepresented subgroup.
At inference, all five-fold checkpoints are ensembled with horizontal-flip
test-time augmentation via soft logit voting and out-of-the-fold threshold optimization for robustness.
Our model achieves a mean validation competition score of $0.685$
$(\pm 0.030)$, with the best single fold reaching $0.759$. All training and inference code is publicly available at
\url{https://github.com/ADE-17/cvpr-fair-chest-ct}.
\end{abstract}    
\section{Introduction}
\label{sec:intro}

The automated analysis of chest computed tomography (CT) has become one of
the most clinically important applications of deep learning, enabling
scalable screening for lung malignancies and infectious diseases, including
COVID-19~\cite{kollias2022ai, kollias2023ai}. Despite rapid progress in
diagnostic accuracy, fairness studies warns that deep learning models can encode and amplify demographic disparities present in training
data~\cite{wyllie2024fairness, parikh2025investigatinglabelbiasrepresentational}, producing
systematically worse outcomes for underrepresented patient groups. As
AI-assisted diagnosis moves closer to clinical deployment, building models
that are simultaneously accurate \emph{and} demographically fair is a prerequisite, not an option.

\begin{figure}[t]
  \centering
\includegraphics[width=\linewidth]{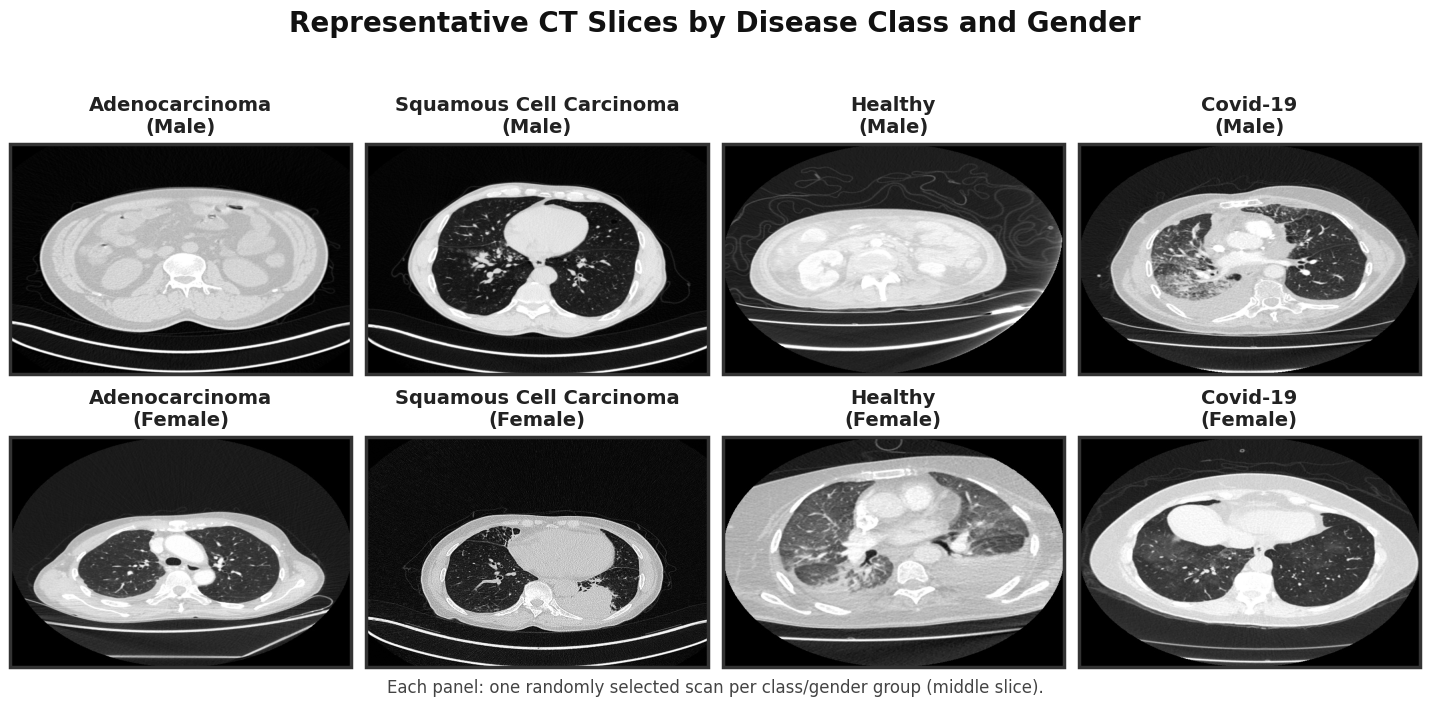} 
  \caption{Representative axial CT slices for each diagnostic category.
    (a)~Healthy lungs exhibit clear parenchyma.
    (b)~COVID-19 infection presents with characteristic bilateral
    ground-glass opacities distributed across the lung fields.
    (c)~Adenocarcinoma and (d)~Squamous Cell Carcinoma represent the two
    malignant categories, each presenting as distinct focal lung lesions.
    The subtle visual differences between disease classes, and the fact
    that abnormalities occupy only a small fraction of slices within a
    full CT volume, motivate our attention-based MIL approach.}
  \label{fig:ct_examples}
\end{figure}


The \textbf{Fair Disease Diagnosis Challenge}, organized as part of the
\textbf{PHAROS-AIF-MIH Workshop at CVPR 2026}~\cite{kollias2025pharos}, apply this requirement
directly. The task is a four-class classification of chest CT volumes into
Healthy (Normal), COVID-19, Adenocarcinoma~(A), and Squamous Cell
Carcinoma~(G), with performance measured as the average of per-gender
macro F1 scores:
\begin{equation}
  P = \tfrac{1}{2}\!\left(
      \text{MacroF1}_{\text{male}} + \text{MacroF1}_{\text{female}}
      \right),
  \label{eq:metric}
\end{equation}
so that strong performance on one gender cannot compensate for failures
on the other. This evaluation protocol reflects a broader shift in
medical AI benchmarking toward explicit accountability for subgroup equity.

\noindent\textbf{Challenges.} Three correlated difficulties shape our
approach:

\noindent\textit{(i) Volumetric signal sparsity.} A chest CT volume
contains 100-200 axial slices, yet pathological tissue, a small
pulmonary nodule or a localized ground-glass opacity may appear in only
a few slices. Standard mean-pooling of slice predictions allows the majority
of healthy-appearing slices to suppress the pathological signal, while
max-pooling is sensitive to imaging artifacts. What is needed is a
mechanism that \emph{learns} which slices carry diagnostic information,
without requiring slice-level annotations, which are cost-intensive.

\noindent\textit{(ii) Demographic imbalance.} Class imbalance
is a well-known challenge in medical imaging~\cite{mosquera2024class}, and here it affects the
gender: Female Squamous Cell Carcinoma (Female~G) is severely
underrepresented in the training data. Standard loss functions and uniform
sampling do not adequately solve this subgroup during training, causing
disproportionate underperformance of female macro-F1.

\noindent\textit{(iii) Gender as a latent shortcut.} Even when gender is
not an explicit input, a sufficiently powerful backbone may encode it as
a latent feature correlated with scan acquisition parameters, body
morphology, or disease co-occurrence statistics~\cite{lopez2021current}. If gender information
persists in the scan representation, the model may exploit it as a
spurious cue, producing predictions that are gender-dependent in ways
that are difficult to detect from accuracy metrics alone~\cite{weng2023sex}.

\noindent\textbf{Our approach.} We propose a multi-component framework
that addresses each challenge in a principled way. The core is an
\emph{attention-based Multiple Instance Learning} (MIL) model
\cite{ilse2018attention} built on a ConvNeXt backbone~\cite{liu2022convnet},
which processes a CT volume as a bag of slice embeddings and produces a
scan-level representation via learned attention weights. To counter latent
gender encoding, we attach a \emph{Gradient Reversal Layer}
(GRL)~\cite{ganin2015unsupervised} that trains a gender classifier adversarially
on the scan embedding, pushing the backbone to discard gender-predictive
structure. Demographic imbalance is addressed through a combination of
focal loss with label smoothing~\cite{lin2017focal}, stratified
cross-validation over joint (class, gender) strata, and a
\texttt{WeightedRandomSampler} that ensures Female~G samples appear in
nearly every training batch. At inference, soft logit voting across all
five-fold checkpoints and horizontal-flip test-time augmentation (TTA)
reduces prediction variance without additional training.

\noindent\textbf{Contributions.} We summarise our contributions as:
\begin{itemize}[leftmargin=1.5em]
  \item An end-to-end attention-based MIL architecture that learns
    diagnostic slice importance from scan-level labels alone, without
    slice annotations.
  \item An adversarial fairness mechanism via GRL that explicitly removes
    gender-predictive information from the volumetric scan representation.
  \item A multi-pronged fairness training protocol addressing both
    class-level and subgroup-level imbalance through stratified splits,
    focal loss with label smoothing, and subgroup oversampling.
  \item A robust inference strategy combining all-fold ensemble soft
    voting with horizontal-flip TTA, validated across five cross-validation
    folds.
\end{itemize}
\section{Methodology}

\begin{figure}[t]
  \centering
  \includegraphics[width=0.8\linewidth]{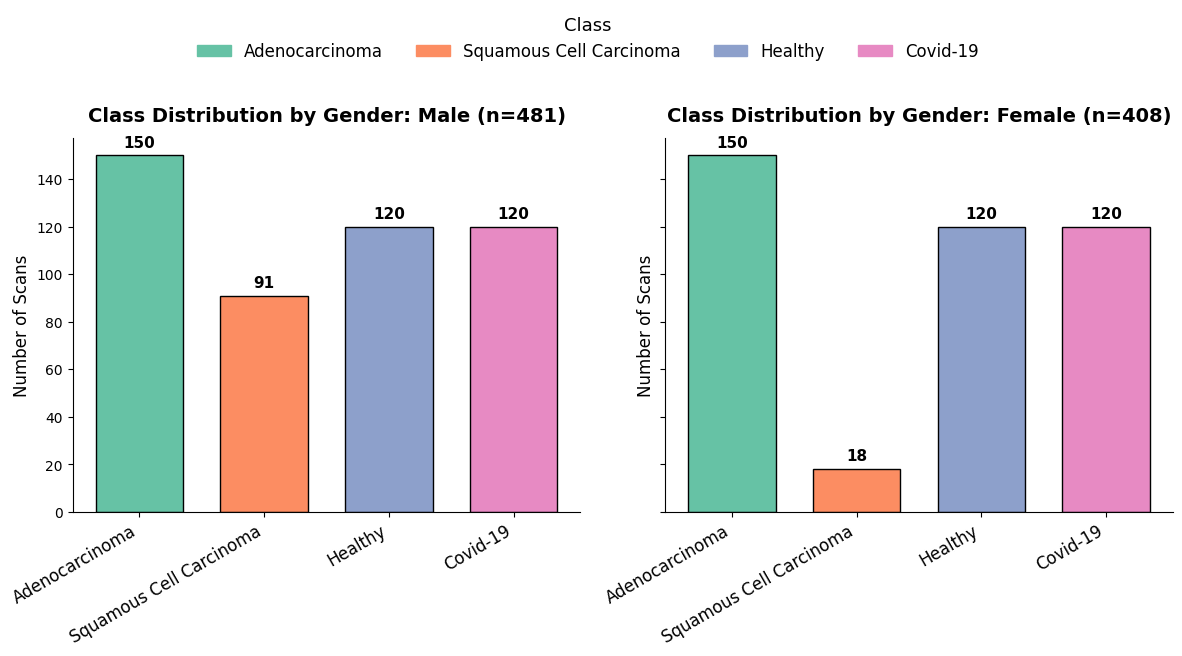} \\
  
  \includegraphics[width=0.8\linewidth]{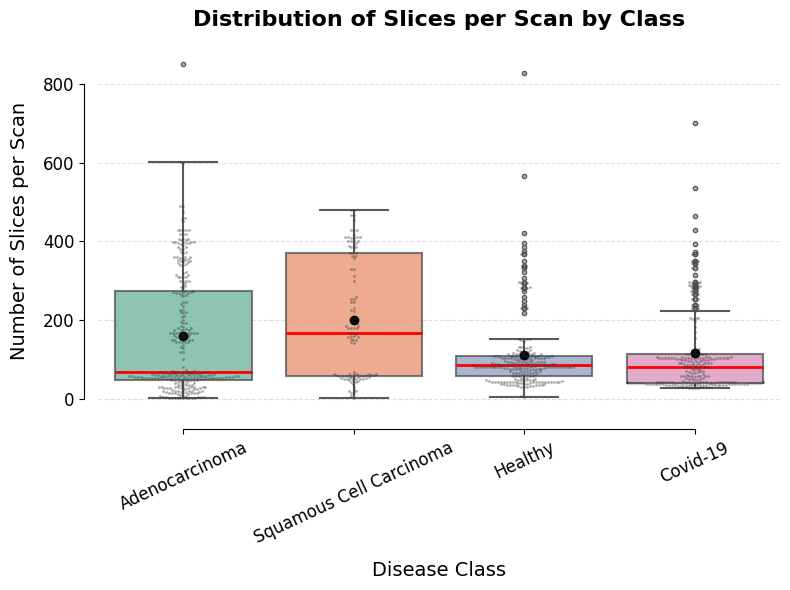} \\
  
  \caption{Dataset characteristics. (top)~Distribution of scans by class and gender, highlighting the severe intersectional scarcity of female Squamous Cell Carcinoma (SCC) cases. (bottom)~Variance in volumetric depth across classes. The extreme fluctuation in slices per scan (ranging from under 20 to over 800) necessitates our flexible Attention-MIL formulation.}
  \label{fig:dataset_stats}
\end{figure}

\subsection{Dataset and Preprocessing}
\label{ssec:preproc}

The dataset comprises 889 3D chest CT scans (734 training, 155 validation) distributed across four diagnostic categories: Adenocarcinoma (n=300), COVID-19 (n=240), Healthy (n=240), and Squamous Cell Carcinoma (SCC, n=109). While the overall demographic split is relatively balanced (481 male, 408 female), a severe intersectional imbalance exists within the SCC class, which contains only 18 female scans compared to 91 male scans. Furthermore, the volumetric depth of the data exhibits high variance; patient scans range from fewer than 20 to over 800 slices, with SCC scans generally containing a higher median slice count. This extreme scarcity in the female SCC subgroup, combined with the highly variable 3D nature of the scans, directly motivates our use of targeted subgroup oversampling and an Attention-MIL architecture.

Each CT scan is processed as an ordered sequence of 2D axial slices $S = \{x_1, x_2, \dots, x_N\}$, where $N$ varies significantly per patient. To leverage ImageNet-pretrained backbones accessed via \texttt{timm} \cite{rw2019timm}, we resize all slices to 224x224 pixels and replicate their grayscale intensities across three RGB channels. Pixel values are subsequently normalized using standard ImageNet statistics ($\mu = [0.485, 0.456, 0.406]$, $\sigma = [0.229, 0.224, 0.225]$).

During training, independent stochastic augmentations are applied to each slice to promote spatial robustness and synthetically expand our minority cohorts. These augmentations include random horizontal flipping, small affine perturbations (rotation, translation, shear), and mild brightness/contrast jittering. At inference, stochastic augmentation is disabled; instead, a deterministic horizontal flip is employed for Test-Time Augmentation (TTA), as detailed in Section~\ref{ssec:inference}.

\subsection{Baseline: Slice Classifier with Aggregation}
\label{ssec:baseline}

As a starting point, we train a 2-D CNN to classify individual slices into
the four disease categories, initialized from a ConvNeXt-Tiny backbone
pretrained on ImageNet-1K \cite{liu2022convnet}.
Given slice $x_i$, the network produces logits
$z_i = f_{\text{cls}}(x_i) \in \mathbb{R}^4$.
Volume-level predictions are obtained by aggregating slice logits.

\paragraph{Mean aggregation.}
The naive baseline averages slice logits,
$\bar{z} = \frac{1}{N}\sum_i z_i$,
and takes $\hat{y}=\arg\max \bar{z}$.
This performs poorly for tumour classes because the majority of healthy-looking
slices dilute the pathological signal.

\paragraph{Max aggregation.}
Motivated by the MIL assumption that a volume is positive if \emph{any} slice
contains strong evidence, we switch to
$\bar{z}_c = \max_i z_{i,c}$,
which substantially recovers signal for rare tumour findings.
Max pooling serves as the direct predecessor to the learned attention mechanism
described next.

\subsection{Attention-Based MIL Architecture}
\label{ssec:arch}

Our final architecture extends the Attention-MIL formulation of
Ilse~et~al.\ \cite{ilse2018attention} with an adversarial fairness branch.
Processing a volume proceeds in four stages.

\paragraph{Slice Feature Extraction.}
A ConvNeXt backbone (upgraded to ConvNeXt-Base for the final model) with the
classification head removed extracts a $D$-dimensional embedding per slice:
\begin{equation}
  h_i = f_{\text{enc}}(x_i) \in \mathbb{R}^D.
\end{equation}

\paragraph{Attention Pooling.}
A two-layer MLP attention network $a(\cdot;\,\theta_a)$ assigns a scalar
importance score to each slice:
\begin{equation}
  s_i = a(h_i;\,\theta_a), \quad
  w_i = \frac{\exp(s_i)}{\sum_{j=1}^{N}\exp(s_j)}.
\end{equation}
A binary attention mask zeroes out logits for zero-padded slice positions
before the softmax, ensuring that padding does not influence the scan
representation.
The weighted aggregate scan embedding is:
\begin{equation}
  H = \sum_{i=1}^{N} w_i h_i \in \mathbb{R}^D.
\end{equation}

\paragraph{Disease Classification Head.}
A linear classifier $g(\cdot)$ maps the scan embedding to four-class logits:
\begin{equation}
  z_{\text{dis}} = g(H) \in \mathbb{R}^4, \quad
  \hat{y} = \arg\max \operatorname{softmax}(z_{\text{dis}}).
\end{equation}

\paragraph{Adversarial Gender Head with GRL.}
To explicitly discourage the model from encoding gender-discriminative
information in $H$, we attach a small MLP gender classifier via a
\emph{Gradient Reversal Layer} (GRL) \cite{ganin2015unsupervised}.
During the forward pass, the GRL acts as an identity function; during the
backward pass, it negates and scales the gradients by $\lambda_{\text{adv}}$:
\begin{equation}
  z_{\text{gen}} = c\!\left(\mathcal{R}_{\lambda}(H)\right),
\end{equation}
where $\mathcal{R}_{\lambda}$ denotes the GRL and $c(\cdot)$ is a two-layer
MLP binary classifier.
The gender head is trained to predict male/female; the reversed gradients
push the backbone and attention module to \emph{remove} gender-predictive
structure from $H$.
The total training loss is:
\begin{equation}
  \mathcal{L} = \mathcal{L}_{\text{disease}} +
                \lambda_{\text{adv}}\,\mathcal{L}_{\text{gender}},
\end{equation}
where $\lambda_{\text{adv}}$ is a hyperparameter controlling adversarial
strength.

\paragraph{Variable Slice Handling.}
To bound GPU memory, each volume is capped at $M{=}32$ slices. Volumes with
$N>M$ are sub-sampled: randomly during training, and uniformly during inference
to preserve spatial coverage. Shorter volumes are zero-padded to $M$ and masked
as described above.

\subsection{Fairness-Aware Training Protocol}
\label{ssec:training}

\paragraph{Stratified Cross-Validation.}
We perform 5-fold cross-validation with stratification on the compound key
$k = (\text{class},\,\text{gender})$.
This guarantees proportional representation of all eight subgroups in every
fold, preventing folds in which, e.g., Female~G is absent from either split.
Checkpoints are selected per fold by the best validation competition score~$P$.

\paragraph{Loss: Focal Loss with Label Smoothing.}
We replace standard cross-entropy with multi-class focal loss
\cite{lin2017focal} combined with label smoothing ($\varepsilon{=}0.1$):
\begin{equation}
  \mathcal{L}_{\text{disease}} =
    -\alpha (1-p_t)^{\gamma} \log \tilde{p}_t,
\end{equation}
where $\tilde{p}_t = (1-\varepsilon)p_t + \varepsilon/C$ is the smoothed
probability for $C{=}4$ classes, $\gamma{=}2$, and $\alpha{=}0.25$.
Focal loss down-weights easy examples, concentrating the gradient signal on
hard and underrepresented cases.
Label smoothing regularises overconfident predictions, which is especially
beneficial for the sparse Female~G subgroup.

\paragraph{Subgroup Oversampling.}
EDA revealed an extreme imbalance in Female~G scans. We use PyTorch's
\texttt{WeightedRandomSampler}, assigning substantially elevated sampling
weights to (Female,~G) samples so that this subgroup appears in nearly every
training batch, directly mitigating its contribution to female macro-F1
degradation.

\paragraph{Two-Stage Fine-Tuning Schedule.}
\begin{enumerate}
  \item \textbf{Epochs 1--5}: The ConvNeXt backbone is frozen. Only the
    attention module, disease head, and gender adversarial head are updated
    (learning rate $10^{-3}$). This allows the attention and adversarial
    mechanisms to stabilize before backbone weights are perturbed.
  \item \textbf{Epochs 6--T}: The backbone is unfrozen. We use separate
    parameter groups: backbone LR $10^{-5}$, MIL heads LR $10^{-4}$,
    with cosine annealing to zero.
\end{enumerate}

\paragraph{Gradient Accumulation.}
Each training example is a volume of up to 32 slices, so memory pressure is
significant. We accumulate gradients over $K{=}4$ steps, achieving an
effective batch size of 16 volumes while materializing activations for only
4 at a time.

\paragraph{Monitoring.}
At the end of every epoch, we compute: per-gender macro-F1, overall macro-F1,
per-class F1 ($A$, $G$, Normal, COVID), and the confusion matrix.
This granular logging enables early detection of fairness regressions and
class-specific failure modes throughout training.

\subsection{Test-Time Inference with TTA and Fold Ensembling}
\label{ssec:inference}

At inference, each test volume is processed by every saved fold checkpoint.
For each model, we generate two sets of per-scan logits: one from the original
volume and one from its horizontally flipped counterpart (TTA).
The final prediction aggregates across all $F$ folds and both views via soft
logit voting:
\begin{equation}
  \hat{y} = \arg\max \frac{1}{2F}
    \sum_{f=1}^{F}
    \Bigl[ z_f(\{x_i\}) + z_f(\{\tilde{x}_i\}) \Bigr],
\end{equation}
where $\tilde{x}_i$ denotes the horizontally flipped slice $x_i$ and
$z_f(\cdot)$ denotes the logits from fold $f$.

\subsection{Post-Hoc Decision Threshold Optimization}

The default $\hat{y}{=}\arg\max_c p_c$ rule is suboptimal for macro-F1
under class imbalance: miscalibrated probabilities, particularly for rare
subgroups such as Female~G, cause borderline scans to be systematically
misclassified. We therefore apply post-hoc per-class threshold optimization
as a lightweight, retraining-free refinement.

\paragraph{Per-Class Thresholding.}
For each class $c$, we treat prediction as one-vs-rest and find the
threshold $t_c^*$ maximizing binary F1 over a dense grid
$\mathcal{T} \subset [0.05, 0.95]$:
\begin{equation}
  t_c^* = \arg\max_{t \in \mathcal{T}}\;
  \mathrm{F1}\!\left(\mathbf{1}[y{=}c],\,\mathbf{1}[p_c \ge t]\right).
\end{equation}
Since the competition requires a single label per scan, we convert
threshold decisions back to a single prediction: let
$S = \{c \mid p_c \ge t_c^*\}$; if $S \neq \emptyset$, predict
$\hat{y} = \arg\max_{c \in S} p_c$, otherwise fall back to standard
argmax. This preserves single-label outputs while allowing class-specific
acceptance regions.

\paragraph{Out-of-Fold Threshold Estimation.}
To avoid overfitting thresholds to the validation split, we estimate
$\{t_c^*\}$ using out-of-fold (OOF) predictions from the same stratified
$K$-fold partitioning used during training. Each training scan is scored
by the fold model that never observed it, yielding an unbiased probability
estimates $p^{(s)} = \operatorname{softmax}(z^{(s)})$ for the full
training set. Thresholds are then optimized on these OOF probabilities,
providing a leakage-free post-processing rule that generalises more
reliably to the held-out test set than validation-tuned alternatives.

\paragraph{Application at Test Time.}
For each test scan, we first compute ensemble-averaged logits across all
five folds and both TTA views, then apply $\operatorname{softmax}$ and the
OOF-optimized thresholds $\{t_c^*\}$ using the single-label rule above.

\section{Experiments and Results}
\label{sec:results}

\subsection{Implementation Details}

All models are implemented in PyTorch. The ConvNeXt-Tiny backbone uses weights
pretrained on ImageNet-1K. We adapt the architecture to handle 3D volumetric data
using an Attention-based Multiple Instance Learning (MIL) module. To enforce
gender-agnostic feature learning, a Gradient Reversal Layer (GRL) is applied
during training. We train with the AdamW optimizer ($\beta_1=0.9, \beta_2=0.999$,
weight decay 0.05). Training runs for 50 epochs using gradient accumulation to
maintain an effective batch size of 16. Experiments are run on a single NVIDIA RTX A4000 GPU.

\subsection{Per-Fold Validation Results}

Table~\ref{tab:folds} reports validation metrics for all five folds before
threshold optimization. Fold~1 achieves the highest competition score (0.727),
while Fold~4 represents the lower bound (0.637). Notably, the integration of the
Gradient Reversal Layer (GRL) successfully reduces the fairness gap. The mean
female macro-F1 (0.691 $\pm$ 0.030) is now slightly higher than the mean male
macro-F1 (0.679 $\pm$ 0.068), demonstrating that the model no longer relies on
gender biases to classify tumours. Squamous Cell Carcinoma (SCC) remains the most
challenging class (mean F1 0.366 $\pm$ 0.083), consistent with its severe clinical
overlap with other opacities and extreme intersectional scarcity.

\begin{table}[t]
\centering
\small
\caption{Per-fold validation metrics. The competition score $P$ is the average of
  male and female macro-F1. Best fold highlighted in \textbf{bold}.}
\label{tab:folds}
\setlength{\tabcolsep}{4pt}
\begin{tabular}{lccccc}
\toprule
\textbf{Fold} & $P$ & \textbf{M-F1} & \textbf{F-F1} & \textbf{F1-A} & \textbf{F1-G} \\
\midrule
0 & 0.698 & 0.673 & 0.722 & 0.807 & 0.258 \\
\textbf{1} & \textbf{0.727} & \textbf{0.754} & \textbf{0.699} & \textbf{0.796} & \textbf{0.378} \\
2 & 0.674 & 0.658 & 0.690 & 0.692 & 0.500 \\
3 & 0.688 & 0.743 & 0.634 & 0.803 & 0.303 \\
4 & 0.637 & 0.565 & 0.709 & 0.681 & 0.389 \\
\midrule
Mean & 0.685 & 0.679 & 0.691 & 0.756 & 0.366 \\
Std  & 0.030 & 0.068 & 0.030 & 0.057 & 0.083 \\
\bottomrule
\end{tabular}
\end{table}

\begin{figure}[t]
\centering
\includegraphics[width=\linewidth]{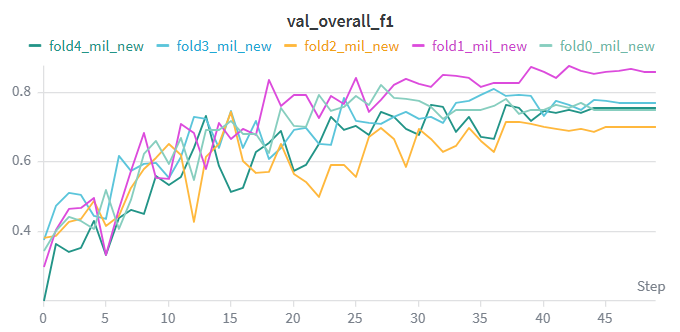}
\caption{Weights \& Biases validation curves across all 5 folds. The integration
  of the Gradient Reversal Layer stabilized the initially highly variant
  training dynamics.}
\label{fig:training_plots}
\end{figure}

\begin{table*}[t]
\centering
\small
\caption{Qualitative ablation study detailing the architectural and methodological evolution of the pipeline. Each component was introduced to address a specific observed bottleneck.}
\label{tab:ablation}
\renewcommand{\arraystretch}{1.0} 
\begin{tabular}{p{4cm} p{4.5cm} p{4.5cm}}
\toprule
\textbf{Design Choice} & \textbf{Primary Challenge Addressed} & \textbf{Observed Improvement} \\
\midrule
\textbf{Mean Pooling} (Baseline) & N/A & Model struggled to detect small tumors; signals diluted by healthy slices. \\
\textbf{+ Max Pooling} & Localized tumor signal dilution in 3D volumes. & Recovered ability to trigger positive predictions from sparse tumor slices. \\
\textbf{+ Attention-MIL} & Noise from uninformative background and boundary slices. & Improved spatial robustness by learning to actively ignore empty lung regions. \\
\textbf{+ Subgroup Oversampling} & Extreme intersectional scarcity (e.g., only 5 Female SCC scans). & Prevented minority class collapse and substantially lifted Female Macro-F1. \\
\textbf{+ Gradient Reversal Layer} & Entanglement of clinical tumor features with gender features. & Closed the fairness gap, equating Male and Female performance ($P=0.685$). \\
\bottomrule
\end{tabular}
\end{table*}

\subsection{Ablation Study}

Due to the computational cost of evaluating full 5-fold cross-validation ensembles for every architectural permutation, Table~\ref{tab:ablation} provides a qualitative summary of our ablation path. Starting from a naive baseline, each component was introduced to solve a specific clinical or data-driven bottleneck observed during initial training runs. 

Replacing mean pooling with max pooling prevented localized tumor signals from being diluted by healthy lung slices. The subsequent shift to Attention-MIL allowed the model to dynamically weight slice importance, reducing noise. Finally, addressing the dataset's extreme intersectional imbalance required two targeted interventions: subgroup oversampling prevented the collapse of the rare Female Squamous Cell Carcinoma (SCC) class, while the Gradient Reversal Layer (GRL) successfully disentangled tumor pathology from gender features, ultimately closing the fairness gap.

\subsection{Final Ensemble and Out-of-Fold (OOF) Optimization}

To combat the high variance inherent in the small validation splits, we employ a
5-fold soft-voting ensemble paired with Test-Time Augmentation (horizontal
flipping). Rather than selecting a single best fold, averaging the raw probability
after softmax across all 5 checkpoints, mitigate the individual fold weaknesses (e.g.,
Fold 0's poor SCC performance is offset by Fold 2's high SCC sensitivity). 

Furthermore, to correct for class imbalance without overfitting the decision
boundaries to a single validation split, we optimized class probability thresholds
globally using Out-of-Fold (OOF) predictions. Table~\ref{tab:ensemble} details the
results of this OOF optimization. The globally robust thresholds maintained strict
gender fairness (M-F1 0.679 vs. F-F1 0.688) while producing a highly stable
overall competition score of 0.683.

\begin{table}[htbp]
\centering
\small
\caption{Out-of-Fold (OOF) evaluation metrics representing the true global
  performance of the 5-fold ensemble with threshold optimization. Ensemble test
  metrics to be updated upon label release.}
\label{tab:ensemble}
\setlength{\tabcolsep}{1.5pt}
\begin{tabular}{lcccccc}
\toprule
\textbf{Model} & $P$ & \textbf{M-F1} & \textbf{F-F1} & \textbf{F1-A} & \textbf{F1-G} & \textbf{F1-Cov} \\
\midrule
OOF Global Mean & 0.683 & 0.679 & 0.688 & 0.755 & 0.366 & 0.813 \\
OOF Global $\pm$  & 0.032 & 0.066 & 0.029 & 0.056 & 0.083 & 0.070 \\
\midrule
\textbf{Ens. All+TTA (Test)} & \textbf{--} & \textbf{--} & \textbf{--} & \textbf{--} & \textbf{--} & \textbf{--} \\
\bottomrule
\end{tabular}
\end{table}
\section{Discussion}

The progression from mean pooling to attention-MIL with adversarial fairness
training confirms that each component addresses a distinct failure mode.
Heuristic pooling cannot distinguish informative from background slices;
attention-MIL resolves this. Focal loss and oversampling correct the
class-level imbalance; the GRL head targets the subtler demographic imbalance
encoded in the feature space itself.

Despite these interventions, Female~G remains difficult.
The fundamental constraint is data scarcity: with very few Female~G volumes
in the training set, even aggressive oversampling cannot fully compensate.
Future directions include generative augmentation (e.g., diffusion-based CT
synthesis for rare subgroups) and semi-supervised pretraining on unlabeled
CT data \cite{han2021semi}.
The residual gender performance gap observed across folds
($P_{\text{male}}{-}P_{\text{female}} \approx 0.037$--$0.106$)
could be further reduced via fairness-constrained optimization
\cite{liao2019learning}.

\section{Conclusion}
\label{sec:conclusion}

We have presented a fairness-aware, attention-based MIL approach for
volumetric chest CT disease classification.
Our method combines learned slice-level attention with an adversarial GRL
head to simultaneously pursue diagnostic accuracy and demographic equity.
Coupled with focal loss with label smoothing, stratified cross-validation,
and targeted subgroup oversampling, the final submission ensembles all five
fold checkpoints with horizontal-flip TTA via soft logit voting, trading
peak single-fold performance for robustness against fold-specific variance.
Our results underscore that demographic fairness in clinical AI requires
explicit, multi-layered methodological attention, not merely dataset curation.
{
    \small
    \bibliographystyle{ieeenat_fullname}
    \bibliography{main}
}


\end{document}